\documentclass[11pt,a4paper]{article}

\usepackage[utf8]{inputenc}
\usepackage[T1]{fontenc}
\usepackage{lmodern}
\usepackage{geometry}
\usepackage{microtype}
\usepackage{graphicx}
\usepackage{booktabs}
\usepackage{longtable}
\usepackage{tabularx}
\usepackage{array}
\usepackage{amsmath}
\usepackage{enumitem}
\usepackage{authblk}
\usepackage[numbers,sort&compress]{natbib}
\usepackage{xcolor}
\usepackage{url}
\usepackage[hidelinks]{hyperref}
\usepackage{bookmark}

\graphicspath{{figures/}}
\setlist{nosep,leftmargin=*}
\newcommand{\tightlist}{\setlength{\itemsep}{0pt}\setlength{\parskip}{0pt}}

\title{MedBeads: An AI-Native Clinical Context Graph Built from\\
Immutable Beads and Reconstructable Clinical Links}
\author[1,2]{Takahito Nakajima}
\affil[1]{Diagnostic Imaging and Interventional Radiology, Institute of Medicine, University of Tsukuba}
\affil[2]{Center for Cyber Medicine Research, University of Tsukuba}
\date{}

\begin{document}
\maketitle

\begin{abstract}
\textbf{Background:}
Generative artificial intelligence can encode substantial medical knowledge, yet its answer to a
patient-specific question is constrained by the context supplied at inference time. Electronic
health records (EHRs) and Fast Healthcare Interoperability Resources (FHIR) are indispensable for
documentation, care delivery, and interoperability, but they do not by themselves define the
complete, current, and auditable context that a generative model should receive. Similarity-based
retrieval can find related text, but it does not guarantee that clinically connected records have
been collected or that omitted context is visible to the model.

\textbf{Objective:}
We introduce MedBeads, an AI-facing clinical record substrate designed to assemble a declared
closure of longitudinal patient information before generation. Its aim is not to make a model
reason, but to make the provenance, completeness boundary, and clinical relationships of the
model's input explicit.

\textbf{Design:}
A \emph{Bead} is an immutable clinical or knowledge object whose identifier is the SHA-256 digest
of its canonical content. Beads are physically appended as frames to patient-scoped \emph{Pod}
files; a Bead ID identifies a logical object, not a separate file. Structural parent edges organize
Beads into a patient-rooted Merkle directed acyclic graph. In contrast, \emph{clinical links} are
typed, patient-scoped relationships derived by signed and versioned knowledge rules. They belong
to a reconstructable interpretation layer and may be recomputed when medical knowledge changes
without rewriting clinical facts. Retrieval follows structural edges and authorized clinical
links, resolves amendments and retractions, and reports policy or token truncation explicitly.
New data trigger projection only for the affected patient; knowledge revisions are rolled out by
patient priority. Separate signature-attestation Beads and clearance policies support a path from
single-hospital to multi-institution operation while keeping content identity, trust, and access
distinct.

\textbf{Reference implementation:}
An open-source Go implementation stores append-only Pods and reconstructable SQLite projections.
File-based conversion of 1,135 synthetic Synthea FHIR patient bundles produced a corpus at
approximately one-million-Bead scale and demonstrated reconstruction of the interpretation layer
and deterministic clinical-link derivation. These experiments establish implementation feasibility
and reproducibility; they do not establish a reduction in clinical hallucination or improved
patient outcomes.

\textbf{Conclusion:}
MedBeads reframes grounding as a data-structure problem: immutable evidence is preserved, changing
medical interpretation is projected, and the context delivered to a generative model is a
policy-bounded, provenance-bearing subgraph rather than an opaque list of similar fragments.
Controlled comparative studies are required to test clinical benefit.
\end{abstract}

\noindent\textbf{Keywords:} electronic health records; generative artificial intelligence;
Merkle DAG; content-addressable storage; graph retrieval; clinical provenance; FHIR; security
clearance; multi-institutional health information exchange

\section{Introduction}\label{1-introduction}

\subsection{Medical knowledge is not patient context}\label{11-medical-knowledge-is-not-patient-context}

Large language models (LLMs) have demonstrated broad medical knowledge, but high benchmark performance does not make their patient-specific output reliable. Human evaluation has continued to identify factual, completeness, and safety errors, and performance on longitudinal EHR tasks deteriorates when the available context is shortened \citep{singhal2023,fleming2023}. The distinction is fundamental: model parameters may contain general medical knowledge, whereas a clinical answer depends on the observations, treatments, decisions, corrections, and constraints that apply to one patient at one time.

A model cannot recover a fact that was not provided merely by being fluent. When the supplied record is partial, however, it may still produce a coherent answer. This creates a dangerous failure mode: missing context and generated completion can become difficult to distinguish. Conventional retrieval-augmented generation (RAG) reduces reliance on parametric memory by retrieving external documents \citep{lewis2020}, but semantic relevance is not equivalent to clinical completeness. A laboratory result may be lexically dissimilar to the medication whose safety it changes; two prescriptions a decade apart may represent longitudinal continuity despite weak textual similarity; and a retracted diagnosis may remain highly similar to the query even though it should not be presented as current.

The central problem addressed in this paper is therefore upstream of the model: \textbf{how should a longitudinal clinical record be organized so that a generative system receives the relevant, current, authorized, and auditable chain of information rather than a convenient sample of fragments?}

\subsection{EHRs and FHIR solve different problems}\label{12-ehrs-and-fhir-solve-different-problems}

EHRs are not merely databases. They support human documentation, orders, billing, audit, and care workflows. FHIR provides a standard resource model and API for exchanging health information, and SMART on FHIR enables applications to operate across heterogeneous EHR products \citep{mandel2016,hl7fhirr4}. These functions should not be replaced by an AI-specific format.

Nevertheless, neither an EHR screen nor a collection of FHIR resources is a ready-made model context. Clinicians interpret the chart using implicit knowledge: they recognize an encounter as an episode, connect a laboratory trend to a medication, distinguish a superseded note from a current one, and notice that a sensitive record should not be disclosed in a particular setting. A generative model requires these relationships to be represented or deterministically assembled. Passing an entire longitudinal chart is often infeasible and can itself degrade attention; selecting a few chunks by similarity can omit decisive evidence. The AI-facing layer must therefore preserve the original record while adding a reproducible way to traverse it.

We call this discrepancy between a documentation-oriented record and the input required for grounded generation the \textbf{clinical context mismatch}.

\begin{figure}[htbp]
  \centering
  \includegraphics[width=\linewidth]{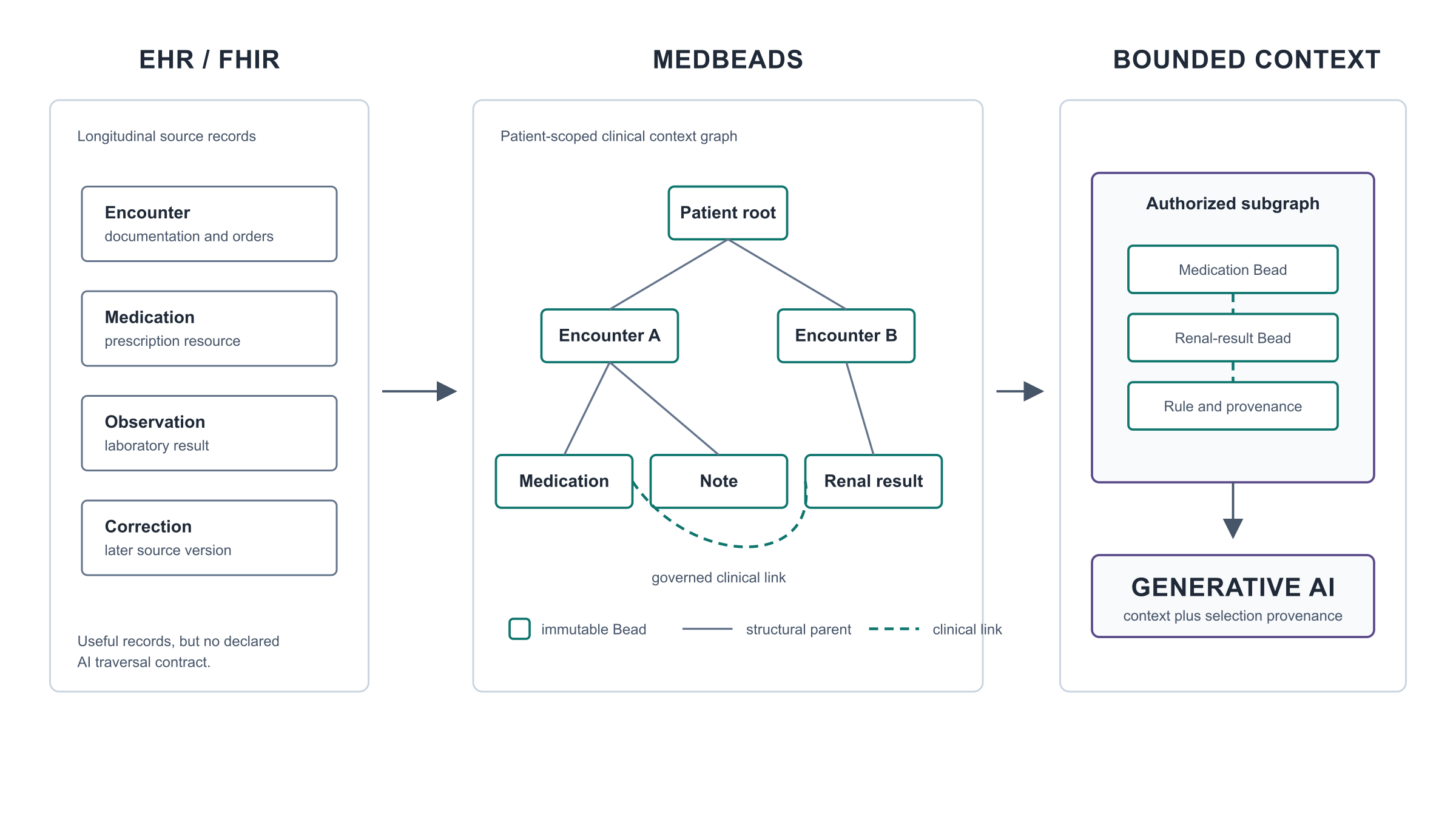}
  \caption{Conceptual flow from fragmented EHR and FHIR inputs through MedBeads to a bounded context supplied to generative AI. Structural parent edges organize immutable Beads; clinical links add governed lateral relationships before retrieval. This is an explanatory illustration and contains no patient data.}
  \label{fig:concept-overview}
\end{figure}

Figure~\ref{fig:concept-overview} summarizes the central transition: MedBeads does not ask the model to infer relationships among isolated records. It assembles a patient-scoped subgraph of Beads and clinical links before generation.

\subsection{The MedBeads hypothesis}\label{13-the-medbeads-hypothesis}

MedBeads is an AI-facing clinical information substrate built around two objects:

\begin{enumerate}
\def\labelenumi{\arabic{enumi}.}
\tightlist
\item
  A \textbf{Bead} is an immutable, content-addressed unit of clinical fact, narrative, correction, provenance, or knowledge.
\item
  A \textbf{clinical link} is a typed relationship between Beads, derived under an explicit knowledge release and stored as a reconstructable projection.
\end{enumerate}

Structural parent edges make the patient record resemble an information tree rooted at the patient and branching through encounters and their events. The structure is technically a directed acyclic graph (DAG), because a Bead may have multiple parents. Clinical links then add lateral relationships across branches and across time. The resulting object is best described as a \textbf{patient-scoped clinical context graph}, not a tree.

This distinction between the node and the edge is the main conceptual contribution. Clinical events should not change when a guideline changes. Their interpretation may. MedBeads therefore preserves clinical and knowledge objects as immutable Beads while allowing clinical links, current-state views, tags, and search structures to be discarded and rebuilt. The system can retain what was recorded, reproduce what an AI was shown under a prior rule generation, and adopt revised knowledge without migrating the fact history.

\subsection{Bounded context completeness}\label{14-bounded-context-completeness}

No software can guarantee that an EHR contains every truth about a patient. MedBeads makes a narrower and testable promise. Given an anchor, an active knowledge release, an authorization policy, and traversal limits, retrieval should return the complete set of reachable, current, authorized Beads under those declared rules---or explicitly report why it did not.

We call this \textbf{bounded context completeness}. It has four implications:

\begin{itemize}
\tightlist
\item
  completeness is relative to a declared graph and policy, not to unknowable clinical reality;
\item
  a token budget is a safety-relevant input, not an invisible implementation detail;
\item
  omitted or truncated references must be enumerated rather than silently dropped; and
\item
  the returned context must carry enough projection provenance to reproduce the relationships that selected it.
\end{itemize}

The hypothesis of MedBeads is that this explicit context contract will make generative output more reproducible and reduce errors caused by missing or stale record context. That hypothesis remains to be tested against conventional FHIR serialization, vector RAG, and graph-based retrieval in controlled LLM studies.

\subsection{Contributions}\label{15-contributions}

This concept and system paper makes six contributions:

\begin{enumerate}
\def\labelenumi{\arabic{enumi}.}
\tightlist
\item
  It defines a two-layer clinical data model separating immutable Beads from reconstructable interpretation.
\item
  It introduces versioned clinical links as auditable, reconfigurable edges for assembling generative-AI context.
\item
  It presents a patient-local incremental projection model and a prioritized rolling strategy for knowledge updates in continuously growing records.
\item
  It separates content identity, digital signature, trust, and security clearance, enabling a single-hospital deployment to evolve toward multi-institution use.
\item
  It positions FHIR as the interoperability boundary and MedBeads as the AI-facing context layer, with source-preserving conversion semantics.
\item
  It reports reference-implementation feasibility while explicitly separating structural validation from the still-unproven clinical effect on LLM hallucination and reasoning.
\end{enumerate}

\section{Conceptual Model: From a Clinical Information Tree to a Context Graph}\label{2-conceptual-model-from-a-clinical-information-tree-to-a-context-graph}

\subsection{Beads are logical objects; Pods are physical storage}\label{21-beads-are-logical-objects-pods-are-physical-storage}

A Bead is the smallest addressable object in MedBeads. It contains a type, clinical time, author identifier, structural parents, correction references, structured or narrative content, and evidence references. The Bead ID is computed from the SHA-256 digest of a JSON Canonicalization Scheme representation \citep{rfc8785}. If any hashed field changes, the identifier changes.

Content addressing provides identity and tamper evidence, but it does not require one file per Bead. Creating millions of tiny files would impose directory, metadata, backup, and locality costs. The current implementation therefore appends Beads as independently framed records to a patient-scoped \textbf{Pod} file. A Pod is a physical pack file; a Bead is a logical immutable object within it. The patient history is read sequentially from one Pod, while the Bead remains individually addressable by its hash through the index.

This resolves a common ambiguity in descriptions of the system:

\begin{quote}
SHA-256 is the Bead ID. It is not necessarily the Bead filename. The immutable content is stored as a frame appended to the patient\textquotesingle s Pod.
\end{quote}

The Pod is append-only, not content-addressed as a whole. Each frame carries corruption detection and enough metadata to recover the logical Bead. If the searchable index is lost, the Beads and structural edges can be reconstructed from the Pods.

MedBeads shares with the InterPlanetary File System (IPFS) the broad use of content addressing and hash-linked Merkle DAG objects \citep{benet2014ipfs}. The architectural purposes differ. IPFS is a general-purpose, versioned peer-to-peer file system for locating and exchanging content-addressed objects. MedBeads neither requires nor currently implements IPFS networking, distributed hash tables, peer discovery, pinning, or CID-based block exchange. Instead, SHA-256 Bead IDs identify canonical clinical objects packed into patient-scoped append-only Pods. Its distinguishing concern is the separation of the immutable structural Merkle DAG from a governed, reconstructable clinical-link layer, together with patient identity, provenance, signatures, correction semantics, and authorization. IPFS could in principle serve as a future storage or replication backend, but it is not part of the MedBeads data model or trust model.

\subsection{The immutable Bead layer}\label{22-the-immutable-bead-layer}

The immutable Bead layer contains the clinical, knowledge, and provenance objects needed to preserve the record and reproduce its interpretation. These objects are not all clinical facts. Immutability describes their identity and history, not their truth or perpetual validity. Examples include observations, encounters, medication requests, conditions, clinical notes, assessments, amendments, retractions, signature attestations, tag dictionaries, and link rules. An erroneous or superseded statement need not remain clinically active: correction occurs by adding a new object that references the prior object rather than rewriting the prior bytes.

Three relations are especially important:

\begin{itemize}
\tightlist
\item
  \texttt{parents} encode structural or provenance dependency;
\item
  \texttt{amends} points from a correction to the record it revises; and
\item
  \texttt{retracts} points from a withdrawal event to the record that must no longer be treated as active.
\end{itemize}

Because a target identifier must exist before it can be referenced, these edges naturally point backward and preserve acyclicity. A deterministic state projection selects the current version while retaining the full history.

Knowledge is likewise stored as versioned Beads. A \texttt{link\_rule} Bead contains the parameters and assertions used to derive clinical links. Publishing a new rule creates a new Bead; it does not overwrite the old rule. A knowledge release closes over an exact set of rule IDs so that an interpretation generation can name every knowledge object it used.

\subsection{The reconstructable interpretation layer}\label{23-the-reconstructable-interpretation-layer}

The interpretation layer contains data that are useful but are not source facts: normalized tags, clinical links, current record status, active condition and medication views, summaries, full-text indexes, and optional vector indexes. These values may change when a mapping improves, a guideline is revised, or projector software is corrected.

The layer obeys the reconstruction function

\texttt{I\_g\ =\ P(F,\ K\_g,\ C\_g,\ V\_g)},

where \texttt{F} is the immutable fact set, \texttt{K\_g} is the closed knowledge set for generation \texttt{g}, \texttt{C\_g} is projection configuration, and \texttt{V\_g} is the projector contract or code version. A projection manifest records these inputs, and projected rows carry a \texttt{projection\_run\_id}. Reproducing an old interpretation therefore requires more than the Pods alone: it requires the Pods plus the frozen knowledge and projection context.

This asymmetric reconstruction guarantee is intentional:

\begin{itemize}
\tightlist
\item
  facts are recoverable from immutable Pods;
\item
  interpretations are reproducible from facts plus a declared knowledge and software generation.
\end{itemize}

The interpretation database is operationally valuable but epistemically disposable. A defect in a summary flattener or link selection algorithm is repaired by re-projecting, not by changing the original clinical events.

\begin{figure}[htbp]
  \centering
  \includegraphics[width=\linewidth]{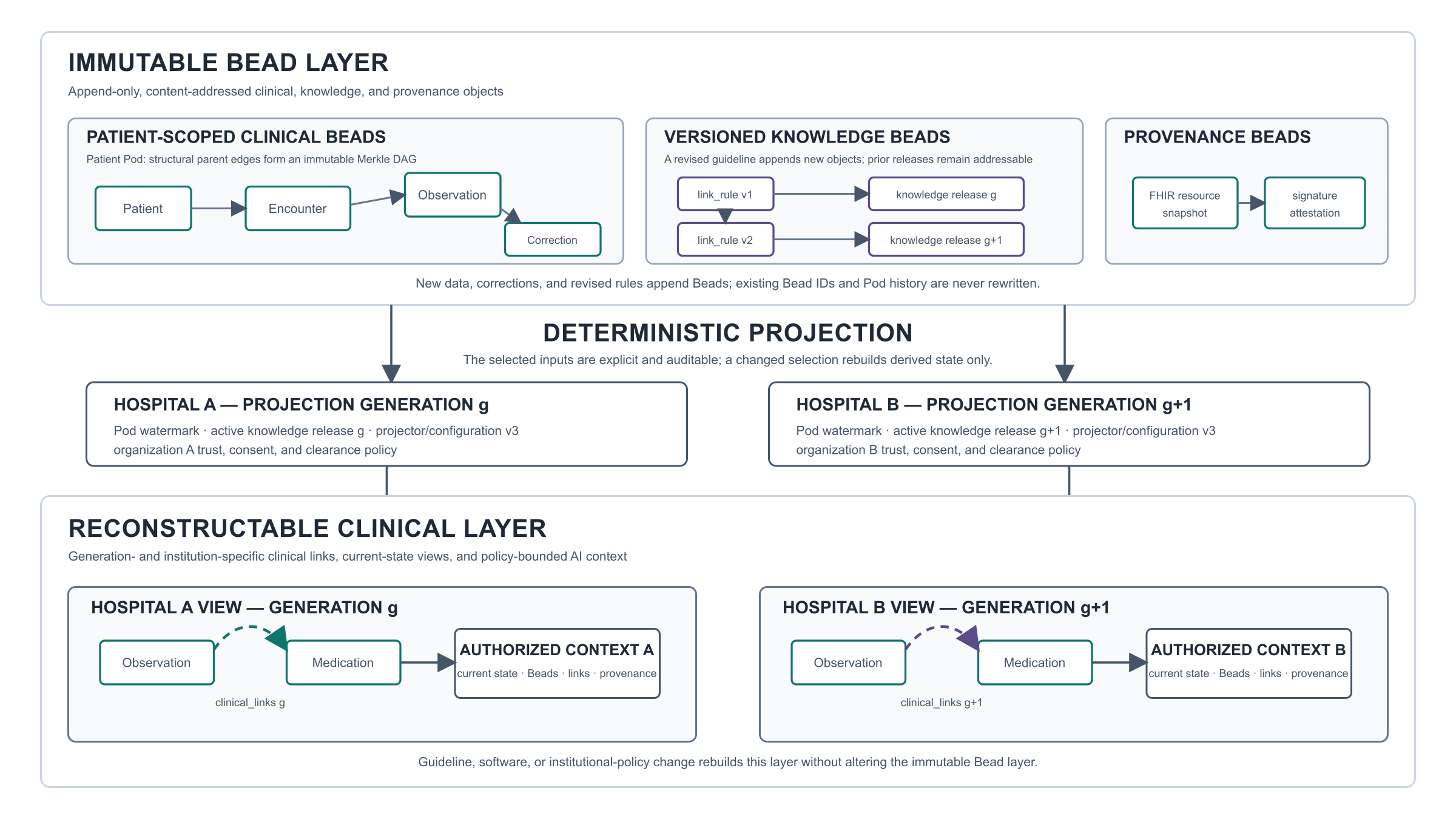}
  \caption{The MedBeads two-layer architecture. Clinical, knowledge, and provenance Beads are append-only, content-addressed objects; corrections or revised guidance create new Beads rather than overwriting existing identities. Deterministic projection records the selected Pod watermark, knowledge release, projector and configuration version, and institutional policy. The same preserved Bead layer can therefore produce generation- or institution-specific clinical links, current-state views, and authorized AI contexts. Rebuilding the clinical layer never rewrites source Beads. This is an explanatory illustration and contains no patient data.}
  \label{fig:architecture}
\end{figure}

Figure~\ref{fig:architecture} illustrates two independent sources of variation. A guideline or software revision changes the projection generation. Multi-institution operation can reuse shared or replicated Beads while each organization selects trusted knowledge and applies its own consent and clearance policy. These choices may change derived clinical links and/or the authorized context, but not the identity or history of the underlying Beads.

\subsection{Structural edges and clinical edges}\label{24-structural-edges-and-clinical-edges}

Parent edges answer questions such as "which encounter contains this observation?" or "which source snapshot produced this mapped clinical object?" They are part of the immutable object contract. Clinical links answer questions such as "which renal measurement is relevant to this medication?", "which orders represent the same therapy over time?", or "which guideline connects these findings?" They depend on clinical knowledge and therefore belong to the interpretation layer.

Conflating these edge classes would force an undesirable choice. If every clinical relationship were immutable, an obsolete guideline would remain embedded in the patient record. If all edges were mutable, provenance and structural history could be silently altered. MedBeads keeps the structural spine immutable and the clinical connective tissue reconstructable.

\begin{table}[htbp]
\centering
\small
\caption{The two edge classes in MedBeads.}
\label{tab:edge-classes}
\begin{tabularx}{\linewidth}{@{}>{\raggedright\arraybackslash}p{0.19\linewidth}XX@{}}
\toprule
Property & Structural parent edge & Clinical link \\
\midrule
Primary meaning & Containment, chronology, source, or correction dependency & Clinically relevant relationship under declared knowledge \\
Layer & Immutable Bead contract & Reconstructable interpretation \\
Changes when & A new Bead is appended & Facts, rule release, configuration, or projector contract changes \\
Provenance & Parent ID hashed into the child Bead & Rule Bead, evidence Beads, and projection run \\
Typical example & Observation belongs to an encounter & Renal result is relevant to a medication \\
Update scope & Append only & Exact per-patient replacement; rolling generation for knowledge change \\
\bottomrule
\end{tabularx}
\end{table}

\subsection{Corrections, narrative, and current state}\label{25-corrections-narrative-and-current-state}

An AI-facing record must contain more than coded resources. Clinician narrative preserves uncertainty, assessment, intent, and reasoning not captured by discrete fields. MedBeads stores free-text notes as first-class Beads while deriving searchable summaries outside the hash contract.

It must also avoid presenting historical statements as current facts. Record-state projection applies a fixed precedence: a valid retraction dominates; an amendment becomes current only under the required clinical attestation; and competing amendments are resolved by recorded time with a deterministic identifier tie-break. The current problem and medication views then combine this record status with the domain status present in the FHIR-derived content. This allows retrieval to exclude retracted records and substitute the current amended version without destroying audit history.

\subsection{What MedBeads does not claim}\label{26-what-medbeads-does-not-claim}

The content hash does not prove that a statement is true, that its author is who the record says, or that the source EHR was correct. It proves only that the addressed canonical content is unchanged. A digital signature can bind an organization and actor statement to that content, but does not grant access. Security clearance can restrict retrieval, but does not itself provide encryption at rest or patient consent. Clinical links express rule-derived relationships, not causal truth unless the supporting rule and evidence justify that interpretation.

Maintaining these distinctions is necessary for a trustworthy system: identity is not authenticity, authenticity is not authorization, authorization is not clinical validity, and structural completeness is not clinical omniscience.

\section{System Architecture and Continuous-Growth Operation}\label{3-system-architecture-and-continuous-growth-operation}

\subsection{Reference architecture}\label{31-reference-architecture}

The reference implementation is a single Go executable containing the Pod store, SQLite projection index, projectors, retrieval engine, command-line tools, REST interface, and Model Context Protocol (MCP) interface. This packaging is an implementation choice rather than a conceptual requirement. The architectural contract is the boundary between immutable facts and derived interpretation.

The write path is:

\begin{enumerate}
\def\labelenumi{\arabic{enumi}.}
\tightlist
\item
  validate and canonicalize an incoming object;
\item
  compute its Bead ID;
\item
  append a framed record to the appropriate patient Pod;
\item
  index the Bead and structural parents;
\item
  update the affected patient\textquotesingle s interpretations; and
\item
  commit the projection watermark.
\end{enumerate}

The read path is:

\begin{enumerate}
\def\labelenumi{\arabic{enumi}.}
\tightlist
\item
  resolve an anchor or query to a patient scope;
\item
  derive current record status;
\item
  traverse structural edges and selected clinical links;
\item
  apply clearance to every candidate and both endpoints of every edge;
\item
  rank and render the authorized context under a token budget; and
\item
  return provenance plus explicit policy and token truncation metadata.
\end{enumerate}

No projected value flows back into a clinical Bead. This one-way dependency makes re-projection safe.

\subsection{Patient-local incremental projection}\label{32-patient-local-incremental-projection}

Longitudinal records grow continuously. A design that rebuilds a corpus-wide graph whenever a new result arrives would be unsuitable for routine care. MedBeads uses the patient Pod as both a locality boundary and a correctness boundary.

When a new Bead is appended, only that patient is projected. The clinical-link set for the patient is replaced atomically rather than merely adding links incident to the new Bead. This is necessary because thresholds and per-Bead caps can make a new event change which older links qualify. Recomputing the affected patient\textquotesingle s small graph preserves exact rule semantics without rebuilding unrelated patients.

The patient projection state records two different dimensions:

\begin{itemize}
\tightlist
\item
  a \textbf{data watermark}, identifying how far the patient\textquotesingle s Pod has been consumed; and
\item
  an \textbf{interpretation generation}, identifying which knowledge and projector contract produced the current projections.
\end{itemize}

If the process stops after the Pod append but before the SQLite commit, the next open compares watermarks, identifies only inconsistent patients, and catches them up. The immutable append is therefore recoverable without treating the index as the source of truth.

\subsection{Knowledge updates as prioritized rolling projection}\label{33-knowledge-updates-as-prioritized-rolling-projection}

A guideline or \texttt{link\_rule} revision can affect many patients. Updating all patients synchronously would create an avoidable operational spike, especially in a design target of 100,000 to 1,000,000 patients. MedBeads registers a new target interpretation generation and migrates patients in small transactions.

The default schedule prioritizes:

\begin{enumerate}
\def\labelenumi{\arabic{enumi}.}
\tightlist
\item
  patients with a new write, who bypass the queue and are projected immediately;
\item
  patients with recent clinical activity, using a configurable three-year threshold;
\item
  patients with long-inactive records; and
\item
  patients marked deceased for scheduling purposes.
\end{enumerate}

Death and inactivity are scheduling hints only. They do not alter clinical truth, visibility, or retention. Unknown legacy status is biased toward the higher-priority group. Failed patients are retained in a retry table with error and attempt state, whereas the normal backlog is virtual: the engine selects patients whose generation watermark differs from the target. This avoids inserting one million queue rows for a one-million-patient deployment.

During a rolling update, each patient\textquotesingle s projected rows name their generation. Retrieval can therefore identify whether a patient is current or awaiting migration. A production policy may choose to allow the prior signed generation temporarily, force immediate projection on access, or deny high-risk use until the target generation is applied. The current implementation provides the generation metadata and the immediate-on-write path; institution-specific clinical policy remains an operator responsibility.

\subsection{Projection manifests and knowledge releases}\label{34-projection-manifests-and-knowledge-releases}

A projection manifest is append-only. It names the projection, configuration hash, projector contract version, and exact knowledge Bead IDs. An active-generation pointer identifies the intended current interpretation, while patient watermarks record actual rollout progress.

For trusted operation, an arbitrary rule cannot become active merely because it exists in storage. A \texttt{knowledge\_release} Bead declares a closed set of \texttt{link\_rule} IDs, revision label, publishing organization, publication time, and applicability interval. The release becomes eligible only when the trust policy\textquotesingle s required \texttt{knowledge\_release} signatures validate. Startup revalidates the active manifest so a revoked, expired, or tampered release fails closed.

This mechanism resembles software release management more than mutable configuration. The rules, approvals, and manifest remain addressable after supersession, enabling reconstruction of what relationships an AI was allowed to traverse at a past time.

\subsection{Context retrieval contract}\label{35-context-retrieval-contract}

The MCP retrieval operation is bounded by graph depth, maximum expanded links, clearance, status policy, and token budget. Clinical-link traversal is patient-scoped and bounded; the implementation defaults to one clinical-link hop and a small maximum expansion, with hard upper limits. Higher-severity, better-evidenced links are preferred over weaker co-occurrences.

The response distinguishes at least two causes of incompleteness:

\begin{itemize}
\tightlist
\item
  \textbf{policy truncation}, when graph depth or link limits stop expansion; and
\item
  \textbf{token truncation}, when selected content cannot fit the rendering budget.
\end{itemize}

The response enumerates affected references. This prevents the model from receiving an apparently complete bundle when the system knows it is partial. Clearance filtering is not reported as a list of hidden clinical facts to unauthorized callers, because the existence of a sensitive record may itself be protected information. Audit roles may receive a different view under explicit policy.

\subsection{Scaling claims and limits}\label{36-scaling-claims-and-limits}

Patient-local append and projection, virtual generation queues, and short per-patient transactions are intended to support ordinary operation at 100,000 patients and a design horizon of approximately 1,000,000. These are architectural targets, not measured capacity claims. The current large synthetic corpus is approximately one thousand patients and one million Beads. Validation at larger patient counts must include write contention, Pod growth, index size, restart recovery, key-management latency, rolling-generation lag, and authorization cost.

\section{Clinical Links: Reconstructable Edges for Generative-AI Context}\label{4-clinical-links-reconstructable-edges-for-generative-ai-context}

\subsection{Why parent edges are insufficient}\label{41-why-parent-edges-are-insufficient}

The immutable parent DAG supplies a reliable structural spine, but many clinically relevant relationships are lateral. A medication may depend on renal function measured during another encounter. A later prescription may continue a treatment begun years earlier. A new finding may activate a guideline whose evidence is distributed across observations, diagnoses, and narrative.

Similarity search may retrieve some of these records, but its selection criterion is semantic proximity. The criterion required here is different: whether a versioned clinical rule says that two Beads should be considered together. MedBeads expresses this second criterion as a clinical link.

\subsection{Link schema}\label{42-link-schema}

A clinical link is a derived assertion containing:

\begin{itemize}
\tightlist
\item
  source and target Bead IDs;
\item
  a typed relation;
\item
  the matched normalized tag or rule assertion;
\item
  severity;
\item
  evidence basis, such as co-occurrence, curated knowledge, or guideline;
\item
  the \texttt{link\_rule} Bead ID (\texttt{rule\_version});
\item
  supporting evidence Bead IDs where required;
\item
  deterministic score components;
\item
  a deterministic link identifier; and
\item
  the \texttt{projection\_run\_id} that produced it.
\end{itemize}

The natural key includes the rule version. If two independent rules support the same pair and relation, both assertions are retained instead of allowing the last rule to overwrite the first. Retrieval may consolidate them for display while preserving each provenance path.

\subsection{Link rules are versioned knowledge}\label{43-link-rules-are-versioned-knowledge}

A \texttt{link\_rule} defines which namespaces and Bead types can match, minimum shared evidence, patient-level frequency thresholds, per-Bead link caps, temporal constraints, score weights, author, visible revision, applicability interval, and optional external evidence Beads. The rule itself is content-addressed.

This design separates two histories:

\begin{itemize}
\tightlist
\item
  the patient\textquotesingle s clinical history, which is never rewritten by a rule change; and
\item
  the institution\textquotesingle s interpretive history, in which one signed knowledge release supersedes another.
\end{itemize}

When guidance changes, the engine projects a new relationship generation from the same Beads. The previous link rows may be operationally replaced, but their manifest and rule Beads remain available for audit and reproduction.

\subsection{Safe escalation}\label{44-safe-escalation}

Not every relationship is a clinical warning. A co-occurrence may be useful for navigation but insufficient for action. MedBeads enforces this distinction in the projection schema: an informational relationship may be based on co-occurrence, but warning-level or higher links must identify curated or guideline knowledge, name a concrete rule version, and cite supporting evidence Beads. Unsupported escalation is rejected at storage time.

This constraint does not establish that every admitted warning is clinically correct. Rule authoring, evidence appraisal, local governance, and prospective validation remain necessary. It does prevent an implementation error from silently relabeling an ungrounded association as an actionable alert.

\subsection{Determinism under caps and ranking}\label{45-determinism-under-caps-and-ranking}

Dense patients can generate combinatorial link candidates. The projector therefore applies thresholds, scores, and per-Bead caps. Caps create a subtle determinism requirement: if candidates are consumed in non-deterministic map order, independent projections may select different links even when final rows are sorted.

MedBeads orders candidates by a complete deterministic key before applying caps and derives link timestamps and IDs from immutable inputs rather than the wall clock. Re-projecting identical facts under the same rule and projector contract should produce identical link IDs, not merely the same count. A corpus rebuild is consequently a test of projector correctness as well as storage recovery.

\subsection{Traversal and bounded expansion}\label{46-traversal-and-bounded-expansion}

Clinical links become useful to generative AI only when retrieval follows them. Starting from an anchor set, MedBeads performs bounded graph expansion over authorized patient-local links. Each newly reached Bead is subjected to record-state resolution and clearance before inclusion. Links whose endpoint is retracted, belongs to another patient, or is not visible to the caller are removed as whole assertions.

This produces a context subgraph containing both the selected Beads and the reasons they were selected. The model can receive, for example, a medication, the renal observation connected to it, the relation type, the rule version, and the knowledge evidence. If expansion stops at a depth or count limit, the response reports policy truncation.

\subsection{Relationship to vector RAG and GraphRAG}\label{47-relationship-to-vector-rag-and-graphrag}

Vector RAG, GraphRAG, and MedBeads solve overlapping but distinct problems. Vector RAG retrieves semantically similar chunks from a non-parametric memory \citep{lewis2020}. GraphRAG constructs an entity graph and community summaries to answer questions over a corpus \citep{edge2024}. Both can be useful within a MedBeads deployment.

MedBeads contributes the clinical record substrate beneath those techniques. Its nodes are content-addressed record objects; its structural edges preserve provenance and chronology; and its clinical links are deterministic assertions under governed knowledge. Vector similarity may be added as another projection, but it is not allowed to silently become provenance. A learned graph extractor may propose candidate links, but promoting them to warning-level clinical relationships requires a governed evidence basis.

The principal maintenance distinction is patient locality. A new record does not require re-embedding or re-clustering the entire corpus to maintain the canonical clinical-link view. It triggers exact re-projection for the affected patient. A knowledge change becomes a prioritized rolling projection rather than a rewrite of patient facts.

\subsection{Clinical links as the central MedBeads abstraction}\label{48-clinical-links-as-the-central-medbeads-abstraction}

The Bead makes a clinical statement stable and addressable. The clinical link makes the statement useful in changing medical context. One without the other is insufficient: immutable isolated objects reproduce fragmentation, while mutable edges without immutable endpoints cannot be audited. The combination is the core MedBeads abstraction for generative AI.

\begin{figure}[htbp]
  \centering
  \includegraphics[width=\linewidth]{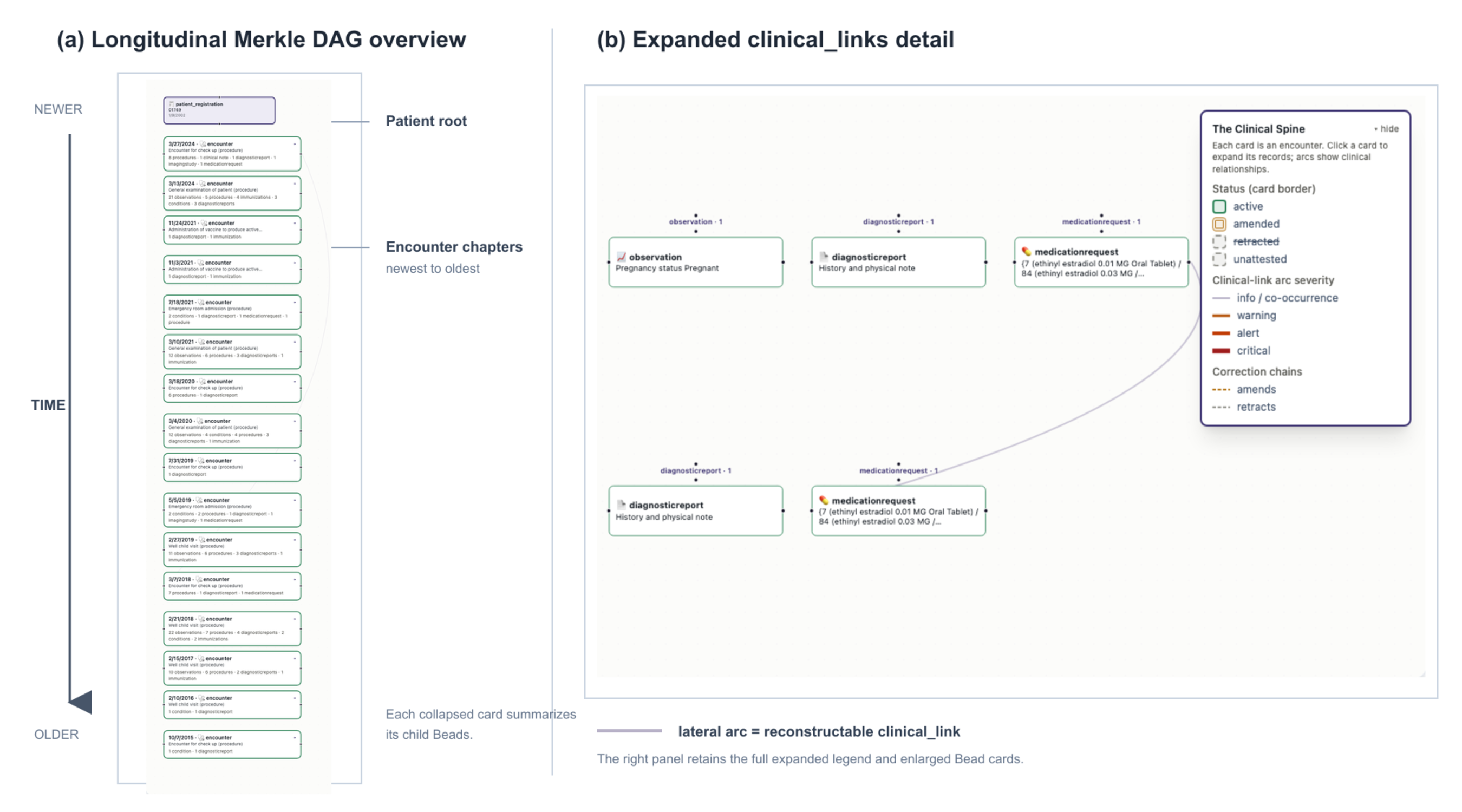}
  \caption{Two scales of the running MedBeads reference interface using bundled synthetic data. (a) A collapsed overview of a longitudinal Merkle DAG, ordered from newer to older, in which each encounter chapter summarizes its child Beads. (b) An expanded detail view in which a curved lateral arc joins two medication-request Beads across encounters; the open legend identifies this arc as a reconstructable clinical link rather than an immutable structural edge. The panels use different synthetic patient examples to preserve legibility at each scale.}
  \label{fig:clinical-links}
\end{figure}

\section{Trust, Security Clearance, FHIR, and Multi-Institution Use}\label{5-trust-security-clearance-fhir-and-multi-institution-use}

\subsection{Hashes and signatures answer different questions}\label{51-hashes-and-signatures-answer-different-questions}

A SHA-256 Bead ID answers: "are these canonical bytes the same object?" It does not answer: "which institution issued this statement?" MedBeads adds the second claim as an immutable \texttt{signature\_attestation} Bead that names its subject as a parent.

The signed statement includes purpose, subject Bead ID, stable organization ID, organization-name snapshot, source-system ID, authenticated actor ID and role, key ID, algorithm, and signing time. It is canonicalized and signed with Ed25519 \citep{rfc8032}. The subject Bead does not change when a signature is attached, and multiple actors or institutions can independently attest the same content.

Three purposes are kept distinct:

\begin{itemize}
\tightlist
\item
  \texttt{clinical\_origin}: the institution asserts clinical origin under its EHR and provenance controls;
\item
  \texttt{fhir\_import}: the connector asserts that it retrieved and transformed a FHIR resource, without claiming that a clinician personally signed it; and
\item
  \texttt{knowledge\_release}: an authorized party approves a closed set of link rules.
\end{itemize}

This avoids a common provenance error: a successful import is not evidence of clinician authorship.

\subsection{Single-hospital deployment}\label{52-single-hospital-deployment}

In an initial single-hospital deployment, clinicians do not need individual private keys. The EHR authenticates the user, and the hospital system signs a statement containing the EHR actor ID, display name at signing, role, source system, and subject Bead ID. The actor is the recorded author; the hospital system is the cryptographic signer.

The local command-line implementation can bootstrap an Ed25519 key for development. A production deployment should place private keys in a hardware security module or key-management service, rotate them under institutional policy, and keep them outside Pods and the SQLite index.

\subsection{Public trust policy}\label{53-public-trust-policy}

Signature verification uses an operator-managed trust policy, never a public key presented only by the attestation itself. The policy binds keys to organizations, accepted purposes, validity intervals, revocation, tenant, and required approval counts. A hospital may trust another institution\textquotesingle s key for \texttt{clinical\_origin} while refusing that key authority to publish local link rules.

Organization and tenant are separate. An organization is the clinical and governance identity that signs statements. A tenant is an operational storage and policy boundary. A shared cloud tenant may host several organizations; one organization may also use several tenants. Stable identifiers drive policy, while human-readable names are display snapshots.

\subsection{Security clearance is not signature trust}\label{54-security-clearance-is-not-signature-trust}

Authenticity does not grant visibility. MedBeads applies a separate clearance policy to Beads and graph traversal. When a clinical link joins an accessible record to a restricted record, the link is removed rather than showing a masked endpoint. This prevents the relationship itself from revealing the existence or nature of sensitive information.

FHIR security labels similarly require an external policy and mutual trust framework to acquire operational meaning \citep{hl7securitylabels}. MedBeads can ingest or map such labels, but local authorization, consent, purpose of use, emergency access, and audit remain institutional responsibilities. The current role-based clearance mechanism is a reference implementation, not a complete legal or consent framework.

\subsection{FHIR as the interoperability boundary}\label{55-fhir-as-the-interoperability-boundary}

FHIR should remain the exchange contract with EHRs. MedBeads should not require hospitals to replace standard APIs or resource semantics. The transformation into Beads serves a different purpose: preserving source versions while organizing an AI-facing graph and its reconstruction provenance.

The proposed server connector separates:

\begin{enumerate}
\def\labelenumi{\arabic{enumi}.}
\tightlist
\item
  an immutable \texttt{fhir\_resource\_snapshot} Bead containing the canonical source resource and source identity; and
\item
  a mapped clinical Bead, such as an observation or medication request, which names the snapshot as a parent and records the mapping version.
\end{enumerate}

This prevents reference resolution, base64 text extraction, code normalization, or other useful transformations from being mistaken for the unmodified source. If mapping logic changes, a new clinical Bead can be generated from the same snapshot.

FHIR logical IDs are only unique within a server namespace. Source identity therefore includes tenant, organization, FHIR server, resource type, logical ID, version ID, last-update time, canonical source digest, and mapping version. The same version key with a different digest is quarantined rather than overwritten.

FHIR R4 history can expose created, updated, and deleted resource versions and supports \texttt{\_since} synchronization \citep{hl7http}. FHIR Provenance can describe creation, transformation, signatures, and version-specific targets \citep{hl7provenance}. These facilities align with MedBeads\textquotesingle{} append-only correction model, but server capabilities vary. A robust connector must preserve deletes as retraction events, distinguish source provenance from importer attestation, and advance a checkpoint only after an entire page has been ingested successfully.

\subsection{Multi-institution records}\label{56-multi-institution-records}

Content identity and patient identity are different questions. A byte-identical clinical object that has already been exchanged without transformation can retain one Bead ID while accumulating independent signature attestations from participating institutions. Independently created records from two hospitals will normally have different Bead IDs because their authors, parents, source identities, or timestamps differ. Neither case proves that the hospitals' patient records concern the same person.

FHIR provides interoperable structures for expressing an identity claim rather than a guarantee that a match is correct. \texttt{Patient.identifier} is a business identifier whose namespace must be retained. \texttt{Patient.link} can assert that another Patient resource concerns the same actual patient, and \texttt{Person.link} can associate records across systems with an identity-assurance level \citep{hl7patientr4,hl7personr4}. The issuer or master-patient-index service remains accountable for the matching process.

The proposed MedBeads path therefore preserves one patient root and Pod per institution. It stores the FHIR identity assertion and provenance as immutable source evidence and converts only an approved assertion into a signed \texttt{patient\_identity\_link} Bead. Candidate matches do not become ordinary \texttt{clinical\_links}. The identity Bead records both roots, issuer, assurance, evidence, adjudication status, effective time, signature, and policy version; a later rejection or correction retracts or supersedes the assertion without merging or rewriting either patient history. A receiving tenant may construct a federated view only after trust, consent, purpose-of-use, and clearance checks.

Cross-institution use nevertheless requires mechanisms outside the Bead hash:

\begin{itemize}
\tightlist
\item
  patient identity must be linked explicitly; \texttt{Patient/\{id\}} cannot be assumed globally unique or compared without its server namespace;
\item
  consent and purpose of use must authorize disclosure;
\item
  organizations must establish trust anchors and key lifecycle policy;
\item
  duplicate or conflicting clinical statements must remain distinguishable; and
\item
  a shared platform must isolate tenants and audit cross-boundary access.
\end{itemize}

MedBeads supplies a structure in which these claims can be attached and audited. It does not make institutional trust or patient consent automatic.

\begin{figure}[htbp]
  \centering
  \includegraphics[width=\linewidth]{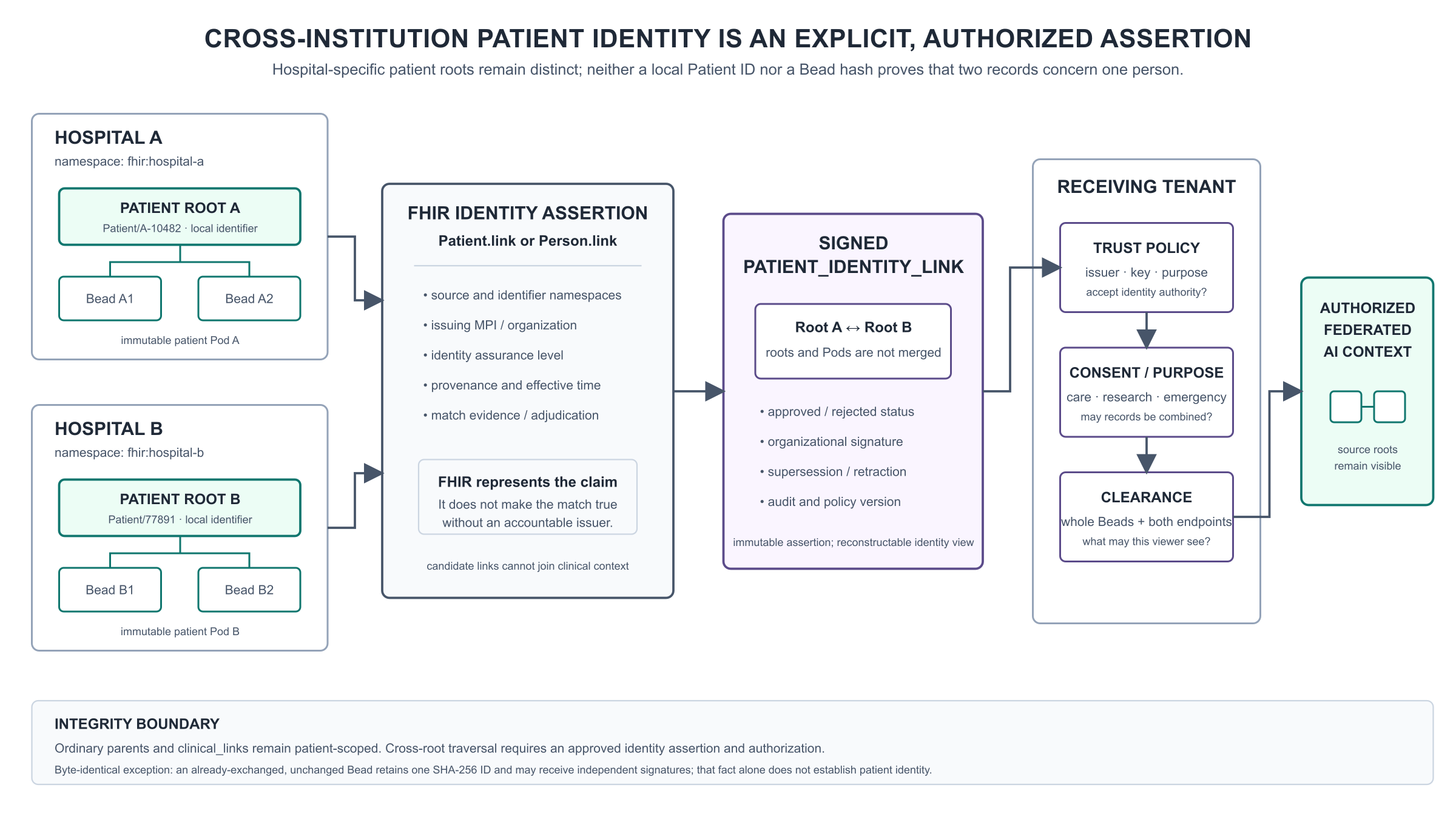}
  \caption{Proposed cross-institution patient-identity and authorization path. Hospital-specific patient roots and Pods remain distinct. A namespaced FHIR \texttt{Patient.link} or \texttt{Person.link} assertion, including issuer, assurance, provenance, and adjudication, may become a signed \texttt{patient\_identity\_link} Bead. The receiving tenant then applies trust, consent and purpose, and security clearance before constructing a federated AI context. A byte-identical Bead may retain one SHA-256 ID after exchange, but content identity alone does not establish patient identity. This is an architectural illustration; cross-institution identity resolution is not implemented in the present reference system.}
  \label{fig:trust-clearance}
\end{figure}

Figure~\ref{fig:trust-clearance} distinguishes an accountable patient-identity assertion from both content identity and access authorization.

\subsection{Preventing cross-patient graph contamination}\label{57-preventing-cross-patient-graph-contamination}

A patient Pod is a physical locality and recovery boundary, but it is not sufficient by itself to prevent every logical cross-patient edge. The current projector derives \texttt{clinical\_links} from tags already filtered by \texttt{patient\_root}, and retrieval fails closed when a link endpoint is outside the loaded patient bundle. Cross-patient amendment and retraction targets are also rejected. However, the reference implementation currently resolves a Bead whose structural parents span multiple roots to the shared Pod, and the SQLite \texttt{clinical\_links} table does not yet have a cross-table trigger proving that both endpoints belong to the row's declared patient root. Complete contamination resistance is therefore a production invariant and evaluation target, not a demonstrated property of the present implementation.

A hardened implementation should enforce the same invariant independently at four boundaries:

\begin{enumerate}
\def\labelenumi{\arabic{enumi}.}
\item
  \textbf{Ingest:} bind every patient-scoped input to an expected root derived from its namespaced FHIR subject; require all patient-scoped parents to share that root; quarantine mismatches instead of moving them to the shared Pod. Shared knowledge is referenced as evidence or governed knowledge, not as a patient parent.
\item
  \textbf{Projection and storage:} require $\operatorname{root}(a)=\operatorname{root}(b)=\texttt{clinical\_link.patient\_root}$; repeat the check in SQLite \texttt{INSERT} and \texttt{UPDATE} triggers so a projector defect cannot persist a contaminated row.
\item
  \textbf{Retrieval and audit:} reject endpoints outside the requested patient bundle, verify clearance at both endpoints, and provide an offline integrity scan over Pod metadata, structural edges, correction targets, and clinical links.
\item
  \textbf{Authorized federation:} allow cross-root traversal only through a trusted, approved \texttt{patient\_identity\_link}, and report the source roots and identity assertion in the returned context. Retraction of the identity assertion removes the federated projection without altering either Pod.
\end{enumerate}

These controls can prevent a technically inconsistent edge between two known roots. They cannot by themselves detect a source EHR or identity service that consistently assigned a record to the wrong human. That semantic error requires namespaced identifiers, MPI assurance, provenance, and adjudication. MedBeads' contribution is to make the decision attributable, reversible, and auditable rather than silently fusing the histories.

\subsection{Applications enabled by the separation}\label{58-applications-enabled-by-the-separation}

Once fact, interpretation, signature, and authorization are separate, several applications become possible without changing Bead identity: role-specific AI context, cross-hospital verification of origin, institution-specific guideline projection, reproducible retrospective audit, research views under restricted clearance, and shared cloud operation with local trust policies. These are architectural possibilities. Only the single-node reference mechanisms described in this paper are currently implemented; multi-institution clinical deployment requires additional governance and security validation.

\section{Reference Implementation and Feasibility Observations}\label{6-reference-implementation-and-feasibility-observations}

\subsection{Implementation}\label{61-implementation}

MedBeads is implemented as an open-source Go program under the Apache-2.0 license. The executable hosts the append-only Pod store, SQLite interpretation index, projectors, integrity verification, REST API, and MCP retrieval tools. Patient Beads are zstd-compressed frames with framing checksums. Searchable fields, parent edges, normalized tags, record status, active views, and clinical links are projected into SQLite.

The implementation includes:

\begin{itemize}
\tightlist
\item
  JCS-based content hashing and Bead verification;
\item
  patient-scoped Pod append and recovery;
\item
  deterministic FHIR Bundle conversion;
\item
  current-state resolution for amendment, retraction, and clinical attestation;
\item
  versioned \texttt{link\_rule} projection with deterministic cap selection;
\item
  bounded clinical-link context expansion;
\item
  patient-local automatic projection and rolling knowledge generations;
\item
  role-based clearance filtering;
\item
  Ed25519 signature attestations, trust policies, and signed knowledge releases; and
\item
  startup validation of trusted active knowledge.
\end{itemize}

The live FHIR server connector, production KMS/HSM integration, and multi-institution patient-identity workflow are design specifications rather than completed reference features.

\subsection{Synthetic corpus conversion}\label{62-synthetic-corpus-conversion}

The principal feasibility corpus consisted of 1,135 synthetic longitudinal patient Bundles generated with Synthea \citep{walonoski2018}. The file-based converter ingested all patients without a failed patient or silent parent fallback. After clinical-note conversion, the patient fact layer contained 961,578 Beads, including one current narrative note per patient. One shared link-rule knowledge Bead brought the addressed store to 961,579 Beads.

This corpus is useful for engineering because it contains longitudinal coded events at approximately one-million-object scale without protected health information. It is not a substitute for clinical validation. Synthea\textquotesingle s disease distribution, documentation, coding, and note-generation behavior are synthetic and materially simpler than production EHR data.

\subsection{Reconstruction and retrieval measurements}\label{63-reconstruction-and-retrieval-measurements}

Measurements were obtained on a single Apple M4 Max workstation with 128 GB memory. The values characterize one implementation and corpus rather than a general performance guarantee.

\begin{table}[htbp]
\centering
\small
\caption{Engineering measurements on the 1,135-patient synthetic corpus.}
\label{tab:engineering}
\begin{tabularx}{\linewidth}{@{}>{\raggedright\arraybackslash}p{0.27\linewidth}>{\raggedleft\arraybackslash}p{0.17\linewidth}X@{}}
\toprule
Operation & Measured result & Interpretation \\
\midrule
Full integrity verification & 6.49 s & CRC and Bead re-hash across the corpus \\
Reindex from Pods & 6 min 20 s & Rebuild searchable/index structures from immutable storage \\
Reproject interpretation & 2 min 04 s & Re-derive links, state, and active views \\
Complete patient-Pod load & 7.8 ms median & Sequential longitudinal read in a fixed sample \\
Warm FTS-to-patient resolution & 19.5 ms median & Search followed by patient scoping \\
\bottomrule
\end{tabularx}
\end{table}

Reconstruction from copied Pods reproduced the Bead, parent-edge, tag, full-text, record-status, active-condition, and active-medication counts. It also exposed an early nondeterminism defect in the clinical-link projector: candidate links were capped while iterating a Go map, so different eligible links could survive across processes. Ordering candidates before cap application corrected the defect, after which repeated projections reproduced link IDs rather than merely link counts. This failure supports the paper\textquotesingle s separation principle: a projection bug was detectable through reconstruction and repair required no mutation of clinical Beads.

\subsection{Clinical-link observations}\label{64-clinical-link-observations}

The corrected projector derived 659,819 informational clinical links across the 1,135-patient corpus under the evaluated co-occurrence rule generation. Every relationship named its rule Bead and projection generation. The initial corpus therefore demonstrates deterministic edge derivation and graph materialization; it does not demonstrate that these links are clinically sufficient or that their use improves an LLM answer.

Record-state projection produced active views containing 9,897 conditions and 2,734 medications. Because the synthetic corpus contained no true correction chains, correction semantics were exercised primarily through unit, integration, and mutation tests rather than prevalence statistics.

\subsection{Continuous-growth regression set}\label{65-continuous-growth-regression-set}

After patient-local projection, rolling generation updates, and bounded link expansion were introduced, a reproducible ten-patient subset was selected from the 1,135 Synthea Bundles. The initial regression ingest produced 4,202 patient Beads, a shared rule Bead, and 492 clinical links with no patient failure. It verified that a new write projects only the affected patient, that recovery catches up watermark mismatches, that rule generations drain in priority order, and that clinical-link expansion remains patient-, state-, and clearance-scoped.

The ten-patient set is intentionally small enough for routine repository tests. It complements but does not replace the earlier full-corpus engineering measurements. The newest signing and rolling-update mechanisms have not yet been benchmarked at the 100,000- or 1,000,000-patient design horizon.

\subsection{What has and has not been established}\label{66-what-has-and-has-not-been-established}

The implementation evidence supports the following claims:

\begin{itemize}
\tightlist
\item
  Beads can be stored as content-addressed logical objects in append-only patient Pods;
\item
  the interpretation index can be reconstructed from the immutable fact layer and frozen projection inputs;
\item
  clinical links can be derived deterministically and replaced without changing clinical facts;
\item
  new writes can be projected at patient scope; and
\item
  signatures and knowledge releases can be validated independently of Bead identity.
\end{itemize}

It does \textbf{not} establish:

\begin{itemize}
\tightlist
\item
  lower hallucination rates than conventional RAG;
\item
  improved diagnostic or treatment accuracy;
\item
  clinical validity of a production rule library;
\item
  production security or regulatory compliance;
\item
  live EHR/FHIR synchronization correctness; or
\item
  operation at one million patients.
\end{itemize}

These distinctions define the evaluation boundary of the present concept and system paper.

\section{Discussion}\label{7-discussion}

\subsection{The record supplied to AI is a clinical artifact}\label{71-the-record-supplied-to-ai-is-a-clinical-artifact}

Generative-AI discussions often treat context preparation as a technical prelude to the model call. In medicine, the selected context is itself a clinical artifact: omission, stale status, unauthorized disclosure, and unsupported relationships can each change the answer. MedBeads makes context construction explicit, versioned, and reproducible.

The proposal does not assume that structure eliminates model hallucination. A model may still misread complete evidence, apply incorrect medical knowledge, or generate an unsupported conclusion. The narrower claim is that missing, stale, or unauditable input should not be accepted as an invisible property of the retrieval pipeline. MedBeads turns those properties into inspectable system state.

\subsection{An AI-facing record, not a replacement EHR}\label{72-an-ai-facing-record-not-a-replacement-ehr}

FHIR and EHR platforms remain responsible for transactional workflows, source documentation, interoperability, and much of the legal record. MedBeads is a derived AI-facing substrate alongside them. Its two-stage FHIR model is important precisely because it avoids claiming that a mapped Bead is the unmodified source resource.

This complementary position also limits migration risk. A hospital can begin with one EHR, one organization identity, and hospital-system signatures. The same object model can later admit multiple source systems or institutions, but cross-organizational trust, consent, identity resolution, and governance must be added deliberately.

\subsection{Why "information tree" is useful but incomplete}\label{73-why-information-tree-is-useful-but-incomplete}

The tree metaphor communicates locality: patient, encounter, and event form a comprehensible hierarchy. Technically, MedBeads uses a Merkle DAG because events may depend on more than one predecessor and corrections reference prior objects. Clinical links then connect branches laterally. Calling the final object a tree would obscure the primary contribution; calling it an unconstrained graph would obscure its immutable patient-rooted spine. \textbf{Clinical context graph} captures both.

\subsection{Update locality and reconstruction semantics}\label{74-update-locality-and-reconstruction-semantics}

Microsoft Research's GraphRAG supports incremental index updates; MedBeads therefore should not be distinguished by claiming that every corpus addition forces a complete GraphRAG rebuild. The distinction is the update contract. Standard GraphRAG maintains derived entities and relationships, communities, community reports, and embeddings over a source corpus \citep{edge2024,microsoftgraphrag2026}. The affected region of an update follows that derived index structure. In MedBeads, a patient-data append has a declared boundary: the new Bead is appended once and the complete clinical-link projection for exactly that patient is replaced atomically. Unrelated patients are not re-projected.

MedBeads does not eliminate reconstruction. A new event can alter which older links survive thresholds or per-Bead caps, so exact patient projection may reconsider that patient's graph. A knowledge-release change may eventually affect many patients, but it advances a named generation through a prioritized queue rather than requiring a synchronous stop-the-world migration.

\begin{table}[htbp]
\centering
\caption{Incremental-update semantics in GraphRAG and MedBeads. The comparison is architectural: GraphRAG can be extended with application-specific controls, whereas MedBeads makes patient-local reconstruction and clinical governance part of the record contract.}
\label{tab:update-semantics}
\footnotesize
\setlength{\tabcolsep}{4pt}
\renewcommand{\arraystretch}{1.12}
\begin{tabularx}{\textwidth}{@{}>{\raggedright\arraybackslash}p{0.18\textwidth}>{\raggedright\arraybackslash}p{0.36\textwidth}>{\raggedright\arraybackslash}X@{}}
\toprule
\textbf{Property} & \textbf{GraphRAG index update} & \textbf{MedBeads reconstruction contract} \\
\midrule
Primary update unit & Source documents and derived entity, relationship, community, report, and embedding index products & One patient projection for a data append; a rolling patient queue for a knowledge change \\
Update boundary & Follows the entity and community structure derived from the corpus & Declared patient scope for \texttt{clinical\_links} \\
Node identity & Entity and community index objects derived from source text & SHA-256 Bead ID over canonical content \\
Edge semantics & Extracted or co-occurrence relationships used for retrieval and summarization & Hashed structural parents plus deterministic clinical links under versioned rules \\
Correction and current state & Defined by the surrounding application & Explicit amendment/retraction relationships and deterministic current-state projection \\
Reproducibility & Requires preservation of source, models, prompts, configuration, and index artifacts & Requires Beads, knowledge release, projector contract, manifest, and projection generation \\
Retrieval boundary & Relevance-ranked context assembled from index products & Authorized reachable closure under declared traversal limits, with truncation reported \\
Authorization & Defined by the surrounding application & Clearance is checked for each Bead and both endpoints of every traversed clinical link \\
\bottomrule
\end{tabularx}
\end{table}

The resulting claim is narrower but stronger: MedBeads does not remove graph maintenance; it makes reconstruction patient-scoped, deterministic, auditable, and schedulable. GraphRAG or vector search can still be added as an optional reconstructable projection above this substrate.

\begin{figure}[htbp]
\centering
\includegraphics[width=\textwidth]{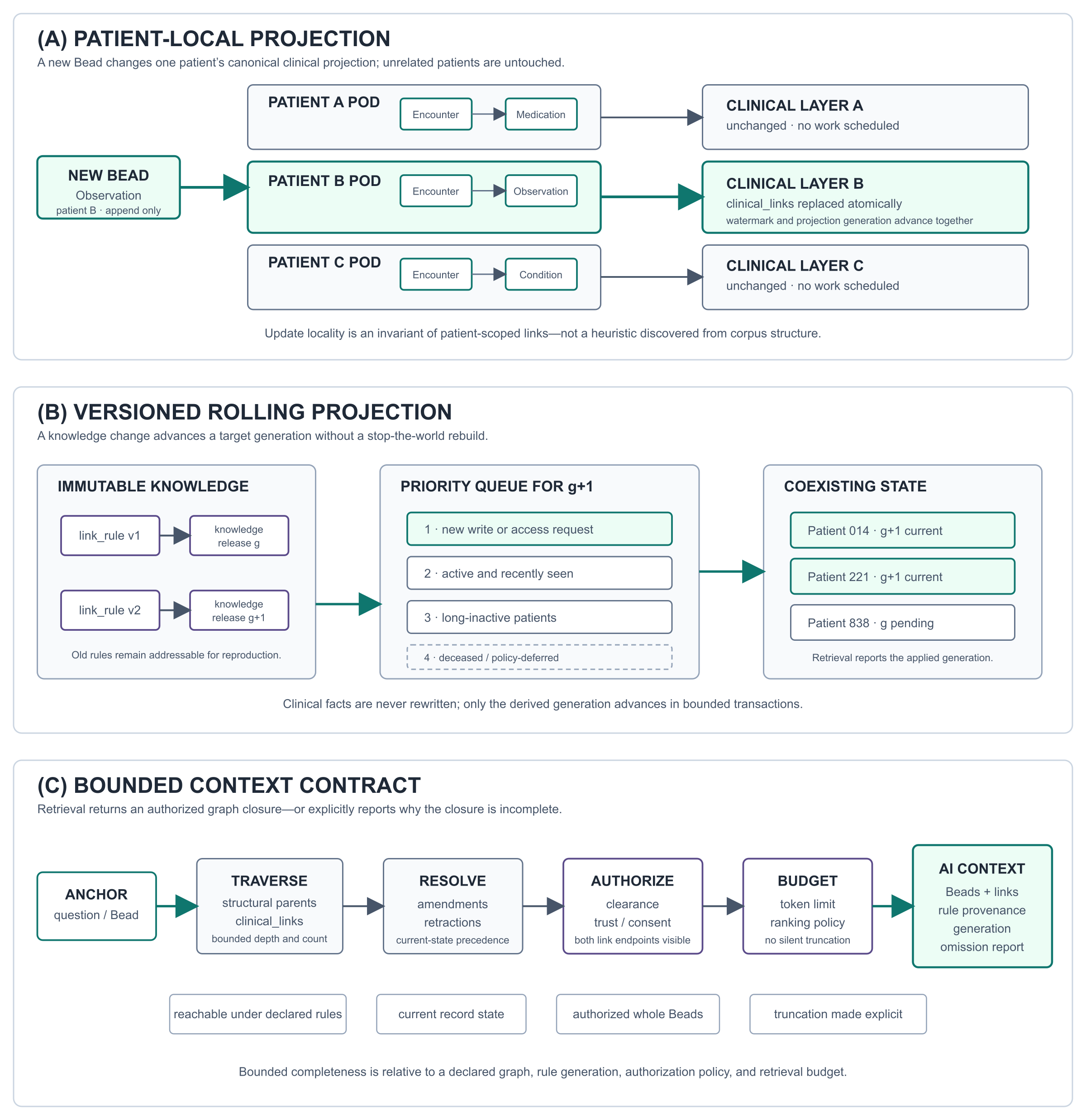}
\caption{Three operational properties of MedBeads. (A) A patient-data append atomically replaces only that patient's clinical projection. (B) A knowledge update advances a named generation through a priority queue while prior generations remain identifiable. (C) Retrieval traverses, resolves, authorizes, and budgets a bounded graph closure and explicitly reports omissions or truncation. The panels describe system contracts and contain no patient data.}
\label{fig:operational-properties}
\end{figure}

\subsection{Reconstructable edges accommodate changing medicine}\label{75-reconstructable-edges-accommodate-changing-medicine}

Clinical facts and clinical interpretation evolve at different rates. A creatinine result recorded in 2025 should not change when a 2027 guideline revises a risk threshold. The relationship between that result and a medication may change. Storing both as equally immutable would either freeze obsolete interpretation or require history rewriting.

Versioned link rules and signed knowledge releases make guideline change a projection operation. Prior interpretations remain reproducible because the facts, rules, approvals, and projector generation are addressable. This pattern may also be useful for terminology mappings, phenotype definitions, active-problem views, and research cohorts.

\subsection{Security and multi-institution implications}\label{76-security-and-multi-institution-implications}

The separation between content identity, patient identity, organizational attestation, trust policy, and clearance is essential for shared records. A common hash can show that two institutions hold the same canonical object, but it cannot show that independently created records concern the same patient. A FHIR identity assertion can express that claim, and independent signatures can show which identity authority or institution vouches for it. Neither fact answers whether a particular clinician, insurer, researcher, patient, or AI agent is permitted to combine or view the records.

This layered model supports institution-specific projections over shared or replicated facts. Hospital A and Hospital B may accept the same signed source object while using different active rule releases, or preserve distinct patient roots connected by an approved identity assertion. They may also apply different clearance or consent decisions. Such flexibility is a benefit only if the boundaries remain visible; collapsing identity, trust, and access into one link would create a serious safety and security error.

\subsection{Limitations}\label{77-limitations}

First, the central clinical hypothesis remains unevaluated. The present experiments measure structural integrity, reconstruction, and systems performance, not hallucination, completeness of clinical reasoning, or patient outcomes. A comparative evaluation should use identical patient questions and model versions across raw FHIR, conventional vector RAG, structural-DAG retrieval, and full MedBeads clinical-link retrieval. Outcomes should include factual support, omission of decisive evidence, temporal correctness, citation fidelity, abstention, and clinician-rated potential harm.

Second, clinical-link quality is bounded by the rule library. The evaluated large corpus used informational co-occurrence links, not a comprehensive curated guideline set. Incorrect or oversensitive rules could create alert fatigue or irrelevant context even when perfectly versioned. Rule authoring requires clinical governance, evidence review, validation cohorts, and monitoring.

Third, the evaluation data are synthetic. Production EHRs contain copy-forward text, malformed references, local codes, duplicated patients, delayed documentation, interface errors, sensitive notes, and institution-specific workflows. Deterministic file conversion does not establish live synchronization correctness.

Fourth, cryptographic and authorization mechanisms are incomplete for production. A local bootstrap key is not an HSM-backed institutional signing service. The current clearance model is not a complete implementation of patient consent, purpose limitation, emergency access, encryption at rest, tenant isolation, or regulatory retention and deletion requirements.

Fifth, the intended 100,000-patient routine scale and approximately 1,000,000-patient design horizon are not benchmarked. Patient-locality reduces the work per new write, but index growth, very large Pods, concurrent ingestion, backup, rolling-generation lag, and operational observability require direct measurement.

Sixth, immutability creates policy questions. Correction is represented well by amendment and retraction, but legal deletion, storage minimization, and key destruction require an explicit retention and cryptographic-erasure design. Tamper evidence should not be confused with a mandate to retain all data indefinitely.

\subsection{Evaluation agenda}\label{78-evaluation-agenda}

A rigorous next research phase should separate four questions:

\begin{enumerate}
\def\labelenumi{\arabic{enumi}.}
\tightlist
\item
  \textbf{Structural completeness:} did retrieval return every authorized current Bead reachable under the declared policy?
\item
  \textbf{Clinical relevance:} did governed clinical links improve recall of decisive evidence without overwhelming context?
\item
  \textbf{Generative performance:} did the same model produce more supported, complete, and temporally correct answers?
\item
  \textbf{Operational safety:} can the system synchronize, recover, rotate knowledge and keys, enforce clearance, and reproduce a past context under realistic load?
\end{enumerate}

These questions require different datasets and metrics. Combining them into one headline accuracy number would hide the mechanism MedBeads is designed to expose.

\subsection{Broader significance}\label{79-broader-significance}

MedBeads suggests that trustworthy medical generative AI may depend as much on the shape and governance of its input as on model scale. The architecture is deliberately model-agnostic. A proprietary cloud model, local model, rules engine, or human reviewer can consume the same provenance-bearing subgraph. This allows the data and context layer to be validated independently of rapid changes in model technology.

\section{Conclusion}\label{8-conclusion}

MedBeads is an AI-facing clinical context graph constructed from two complementary elements: immutable, content-addressed \textbf{Beads} and reconstructable, versioned \textbf{clinical links}. Beads preserve what was recorded and the knowledge objects used to interpret it. Patient-scoped Pods make those logical objects practical to append and read at longitudinal scale. Clinical links connect evidence across events and encounters while remaining replaceable when knowledge changes.

The architecture treats context assembly as a governed clinical operation. Retrieval resolves correction state, follows structural and clinical edges, applies security clearance, respects a token budget, and exposes truncation and projection provenance. Patient-local projection handles routine growth, while prioritized rolling generations distribute the cost of knowledge revision. Separate signature attestations distinguish content identity from organizational origin and create a path from a single-hospital deployment toward multi-institution trust without confusing trust with access.

The reference implementation demonstrates that the two-layer model is technically feasible on a synthetic corpus at approximately one-million-Bead scale and that derived interpretation can be reconstructed without modifying facts. It does not yet prove that MedBeads reduces hallucination or improves clinical decisions. That claim should be tested, not assumed.

The main contribution is therefore a precise change in where responsibility is placed. A generative model should not be asked to infer the shape, currentness, and provenance of an incomplete chart from loosely retrieved fragments. MedBeads prepares a policy-bounded clinical subgraph in which the evidence is immutable, the relationships are governed and reproducible, and any known incompleteness is explicit. This provides a concrete substrate on which trustworthy generative-AI evaluation and future clinical applications can be built.

\bibliographystyle{plain}
\bibliography{references}

\end{document}